\documentclass{article}
\usepackage[citecolor=blue,linkcolor=blue]{hyperref}
\hypersetup{
    colorlinks,
    linkcolor={blue!65!black},
    citecolor={blue!65!black},
    urlcolor={blue!65!black}
}

\usepackage{graphicx}
\usepackage{caption}
\usepackage{amsfonts}
\usepackage{amssymb}
\usepackage{amsthm}
\usepackage{amsmath}
\usepackage{multirow}
\usepackage{algorithmicx}
\usepackage[noend]{algpseudocode}
\usepackage{stmaryrd}
\SetSymbolFont{stmry}{bold}{U}{stmry}{m}{n}
\usepackage{forest}
\usepackage[ruled,vlined,linesnumbered]{algorithm2e}
\usepackage{subcaption}
\usepackage{import}
\usepackage{tikz}
\usepackage{xcolor}
\usepackage{enumerate}
\usepackage{float}
\usepackage{booktabs}

\newcommand{\ud}{\triangleq}
\renewcommand{\vec}[1]{\ensuremath{\boldsymbol{#1}}}

\DeclareMathOperator{\splitt}{\mathit{split}}
\newcommand{\wpp}{\wp_{{\text{\tiny +}}}}

\newcommand{\norm}[1]{\left\lVert#1\right\rVert}

\newcommand{\cH}{\mathcal{H}}

\newcommand{\cL}{\mathcal{L}}

\newcommand{\bR}{\mathbb{R}}
\newcommand{\bN}{\mathbb{N}}
\newcommand{\HR}{\textsc{HR}}

\newcommand{\ra}{\rightarrow}

\newcommand{\true}{\mathbf{t}}
\newcommand{\false}{\mathbf{f}}
\newcommand{\our}{\textrm{MS}}
\newcommand{\ours}{{\textrm{MSs}}}
\newcommand{\fn}[2]{\ensuremath{#1\!\cdot\!10^{#2}}}

\algnewcommand{\IfThenElse}[3]{
  \State \algorithmicif\ #1\ \algorithmicthen\ #2\ \algorithmicelse\ #3}
\algnewcommand{\lIf}[2]{
  \State \algorithmicif\ #1\ \algorithmicthen\ #2}

 \theoremstyle{plain}
 \newtheorem{theorem}{Theorem}[section]

 \theoremstyle{definition}

 \newtheorem{example}[theorem]{Example}

 \theoremstyle{remark}

 \numberwithin{equation}{section}
 
 \theoremstyle{remark}

\begin{document}

\title{Genetic Adversarial Training of Decision Trees}
\author{Francesco Ranzato and Marco Zanella\\
{\normalsize Dipartimento di Matematica, University of Padova, Italy}}
\date{}

\maketitle

\begin{abstract}
We put forward a novel learning methodology
for ensembles of decision trees  based on a genetic algorithm
which is able to train a decision tree for maximizing both its accuracy and
its robustness to adversarial perturbations. This learning algorithm internally leverages on 
a complete formal verification technique for robustness properties of
decision trees based on abstract interpretation, a well known
static program analysis technique. 
We implemented this genetic adversarial training algorithm 
in a tool called Meta-Silvae (\our{}) and we 
experimentally evaluated it on some reference datasets used  
in adversarial training. The experimental results show that 
\our{} is able to train robust models which compete with and often improve on the current state-of-the-art of adversarial training of decision trees while being much
more compact and therefore interpretable and efficient tree models.  
\end{abstract}

\section{Introduction}
\label{sec:introduction}

Adversarial machine learning \cite{cacm18,aml-scale} is a hot topic studying vulnerabilities of machine learning (ML) models in adversarial scenarios. Adversarial examples have been found in diverse application fields of ML, ranging from image classification to malware detection, and the current defense techniques include adversarial model training, input validation, testing and automatic verification of learning algorithms. A ML classifier is defined to be robust for some (typically very small) perturbation of its input samples, which represents an adversarial attack, when it assigns the same class to all the samples within that perturbation, so that unnoticeable malicious alterations of input objects should not deceive a robust classifier. 

This work focuses on the robustness of ML classifiers consisting of decision tree ensembles, such as random forests and gradient boosted decision trees, which are well known for being both accurate and interpretable ML models and are widely used in adversarial scenarios. It has been  amply shown that decision trees can be very non-robust \cite[Section~8.1.4]{James2013}, although it is only recently that robustness verification and adversarial training of tree models  started to be an active subject of investigation \cite{Andriushchenko19,calzavara20,ChenZBH19,Chen2019,EinzigerGSS19,RZ20,tornblom2019,TornblomN20}.

\subsection{Main Contributions}
Genetic algorithms (GAs) \cite{holland1984genetic,Srinivas} provide a widespread effective search technique which computes the next set of hypotheses by repeatedly mutating and then combining parts of the best currently known hypotheses. A number of successful ML methodologies for decision trees are 
based on GAs \cite{BarrosBCF12,Ersoy,Fu,Jankowski,Papagelis,Turney}. Recently, some GAs have also been investigated for adversarial training of neural networks \cite{ChoK19,VidnerovaN20}. To the best of our knowledge, the use of GAs for adversarial training of ensembles of decision trees is still an unexplored topic. In this work we design and experimentally evaluate an adversarial training algorithm for decision tree ensembles based on a genetic algorithm, that we called Meta-Silvae (\our{}, acronym of \emph{Magister Efficiens Temperat Arbores Silvae}, Latin for ``\emph{the efficient master mixes the trees of the forest}''). 
and aims at maximizing both accuracy and robustness of decision trees. \our{} relies on an open source verification method of the robustness of ensembles of decision trees called silva~\cite{RZ20}. This robustness verification algorithm silva performs an abstract interpretation-based static analysis \cite{CC77,rival-yi} of a decision tree classifier which is able to abstractly compute the exact set of leaves of a decision tree which are reachable from an adversarial region. By exploiting this robustness information provided by silva, \our{} is designed as a genetic algorithm that maximizes an objective performance function which is a linear combination of accuracy and robustness.
\our{} is based on some well established design choices of GAs: (1)~elitist selection strategy; (2)~roulette wheel selection; (3)~single-point crossover; (4)~offspring mutation. \our{} has been implemented in C and experimentally evaluated on the reference datasets for adversarial training of decision tree ensembles.  \our{} has been compared with random forests \cite{breiman-random-forests} and with the current state-of-the-art of adversarially trained gradient boosted decision trees \cite{Andriushchenko19,ChenZBH19}. Overall, the experimental results show that \our{} trained models significantly increase their robustness over natural random forests on average of, resp., 3.4$\times$ (with an average absolute gain of $+54.3\%$) while at the same time preserving a comparable accuracy with an expected slight drop (on average $-1.8\%$). Moreover, \our{} models compete with and often improve on the state-of-the-art of adversarially trained gradient boosted decision trees while being much more compact and therefore interpretable (as advocated by \cite{Murdoch} for all ML models) and efficient (as advocated by 
\cite[Section~3.1]{efficient} for ensemble models) tree models.

\begin{figure}[t]
 \centering
 \includegraphics[width=0.45\textwidth]{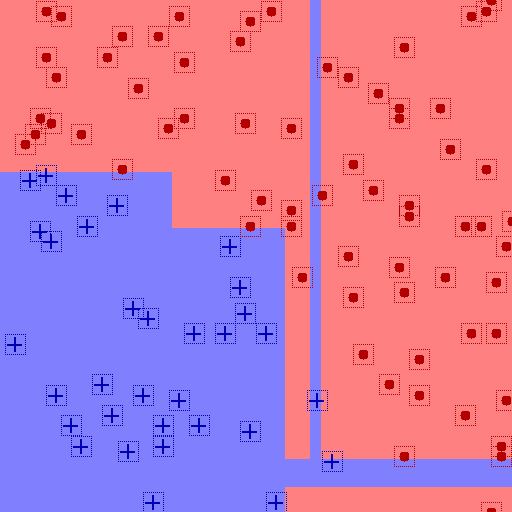}
 \qquad
 \includegraphics[width=0.45\textwidth]{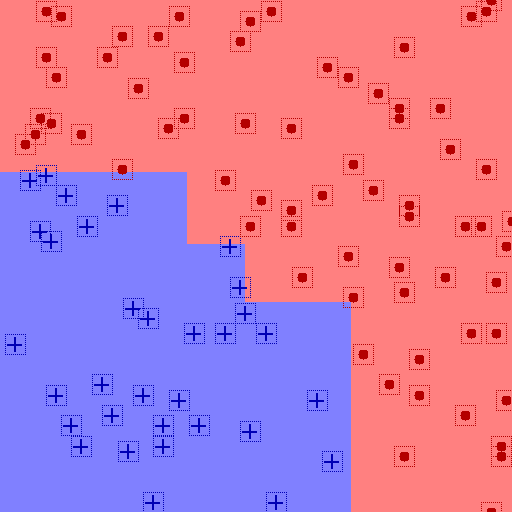}
 \caption{Scikit-learn training (left) vs.\ \our{} training (right)}
 \label{fig:train-standard-vs-our}
\end{figure}

\subsection{Illustrative Example}
Fig.~\ref{fig:train-standard-vs-our} depicts an example of an artificial dataset consisting of 100  2-dimensional random samples $(x_1,x_2)\in[0, 1]^2$ labeled as a blue cross when $x_1^2 + x_2^2 < 0.5$ holds and as a red bullet otherwise. The left diagram represents a decision tree classifier which has been trained by  scikit-learn, while the right diagram has been trained by \our{} using as objective performance function $\varphi($accuracy,\,robustness$)=$ 90\%\,accuracy + 10\%\,robustness. The two diagrams in Fig.~\ref{fig:train-standard-vs-our} show that both classifiers achieve $100\%$ accuracy. The scikit-learn tree  introduces two blue regions cutting the red area in order to achieve an accurate classification of  two blue cross samples, thus making the classification on some samples close to the borders  not robust for a 2$\%$ square perturbation surrounding each input sample. The \our{} training algorithm is able to avoid this lack of robustness since it searches for cut hyperplanes which preserve robustness whenever possible. As displayed by the two diagrams, robustness with respect to the 2$\%$ square perturbation turns out be, resp., $87\%$ and $94\%$ for the left and right decision trees, thus showing a significant increase while still preserving the same 100\% accuracy.

\subsection{Related Work}
While adversarial training of neural networks has been widely studied, few works addressed how to train decision trees which are robust to adversarial attacks \cite{Andriushchenko19,calzavara2019B,calzavara20,ChenZBH19}. In particular, we compared our experimental results with the robust gradient boosted decision trees of \cite{Andriushchenko19,ChenZBH19}. Abstract interpretation \cite{CC77,rival-yi} techniques have been fruitfully applied for designing precise and scalable robustness verification algorithms and adversarial training techniques for a range of ML models \cite{ferrara,vechev-sp18,vechev-icml18,singh-arxiv20,RZ19,RZ20,vechev-nips18,singh2019,SinghGPV-iclr19}. In particular, to our knowledge, \cite{vechev-icml18} is the only work using an abstract interpretation technique for adversarial training of ML models, notably deep neural networks.

\section{Background}
\label{sec:background}

\subsection{Classifiers and Metrics}
Given an input  space $X \subseteq \mathbb{R}^d$ of numerical vectors and a finite set of labels/classes $\mathcal{L} = \{ y_1, \ldots, y_m\}$, a classifier is a function $C : X \rightarrow \wpp(\mathcal{L})$, where  $\wpp(\mathcal{L})$ denotes the set of nonempty subsets of $\mathcal{L}$.
$C$ associates at least one label to every input in $X$ and multiple output labels can be used to model ties in output classification (e.g.\ ties in voting schemes). Training algorithms take a ground truth dataset $D \subseteq  X\times\mathcal{L}$ as input and output a classifier $C: X \rightarrow \wpp(\mathcal{L})$ which minimizes/maximizes some criteria such as a loss function for neural networks or  information gain for decision trees.

Classifiers can be evaluated and compared on the basis of several performance metrics. Accuracy on a test set is a standard metric for assessing a classification model:  given a test set $T \subseteq X\times\mathcal{L}$ of correctly labeled samples, the accuracy of a classifier $C: X \rightarrow \wpp(\mathcal{L})$ on $T$ is  $\mathit{acc}_T(C) \ud |\{(\vec{x},y) \in T ~|~ C(\vec{x}) = \{y\} \}|/|T|$. One typically aims at training classifiers  having a nearly perfect accuracy on suitably crafted test sets. However, according to a growing belief~\cite{cacm18}, accuracy is not enough in ML, because the robustness properties of a classifier may affect its safety and generalization. Given a perturbation function $P: X \rightarrow \wp(X)$ modeling a notion of closeness for input samples, the robustness of $C$  w.r.t.\ $P$ for a test set $T$ is defined by $\mathit{rob}_{T, P}(C) \ud |\{(\vec{x},y) \in T ~|~ C(\vec{x})=\{y\},\: \forall \vec{x'} \in P(\vec{x})\colon C(\vec{x'}) = C(\vec{x}) \}|/|T|$. Perturbation regions $P(\vec{x})\subseteq X$ are used to model adversarial attacks to input samples $\vec{x}$,  i.e., negligible alterations of input vectors aimed at deceiving a classifier. Widely studied perturbations are those induced by $\ell_p$ norms, in particular the $\ell_\infty$ maximum norm \cite{carlini}, which, given an alteration threshold $\epsilon>0$, defines a perturbation $P_{\infty, \epsilon}: X \rightarrow \wp(X)$ as follows: $P_{\infty, \epsilon}(\vec{x}) \ud \{ \vec{x'} \in X ~|~ \norm{\vec{x} - \vec{x'}}_\infty \leq \epsilon \}$, where $\norm{(\vec{x}_1, \ldots, \vec{x}_d)}_\infty = \max\{\vec{x}_1, \ldots, \vec{x}_d\}$. We also consider a more general definition of robustness called stability which encodes the ability of a classifier $C$ of consistently producing the same output on perturbations of input samples, therefore including the cases where $C$ is inaccurate on some input: this is defined as $\mathit{st}_{T, P}(C) \ud |\{(\vec{x},y) \in T ~|~ \forall \vec{x'} \in P(\vec{x})\colon C(\vec{x'}) = C(\vec{x}) \}|/|T|$. We will mostly use stability rather than robustness since we deem stability to be a more comprehensive metric to use in adversarial training.

\subsection{Decision Trees and Tree Ensembles}
Decision trees are well established ML models used for both classification and regression tasks. In this work we consider standard classification trees commonly known as CART (Classification and Regression Trees) \cite{BreimanFOS84}. A classification decision tree $t:X\ra \wpp(\cL)$ is inductively defined as follows. (1)~A base tree $t$ is a leaf $\lambda$ storing a frequency distribution of labels for the samples of the training set which some algorithmic rule (canonically the maximum frequency) converts to one or more predicted labels; thus, we simply consider $\lambda\in \wpp(\mathcal{L})$ and therefore,
for all $\vec{x}\in X$, $t(\vec{x}) \ud \lambda$. (2)~A composite tree $t$  is $\gamma(\splitt,t_l, t_r)$ where $\splitt: X\ra \{\true,\false\}$ is a Boolean split criterion for the internal parent node of its left and right subtrees $t_l$ and $t_r$; thus, for all $\vec{x}\in X$, $t(\vec{x})\ud \textbf{if~} \splitt(\vec{x}) \textbf{~then~} t_l(\vec{x}) \textbf{~else~} t_r(\vec{x})$. A tree stump is of the form $\gamma(\splitt,\lambda_l, \lambda_r)$, i.e., it has a single internal node and two leaves. The training of a decision tree $t$ guarantees that every leaf of $t$ is reachable from at least one sample in the training set. Although split rules in general could be of any type, the most common decision trees employ univariate hard splits of the form  $\splitt(\vec{x}) \ud \vec{x}_i \leq k$ for some feature $i\in [1,d]$ and threshold $k\in \bR$.

Tree ensembles, also known as forests, are sets of decision trees which together contribute to formulate a unique classification output. Training algorithms as well as methods for computing the final output class(es) vary among different tree ensemble models. Random forests (RF) \cite{breiman-random-forests} are a major instance of tree ensemble where each tree of the ensemble is  trained  independently from the other trees on a random subset of the features. Gradient boosted decision trees (GBDT) \cite{friedman2001greedy} represent a different training algorithm where an ensemble of trees is incrementally build by training each new tree on the basis of the data samples which are mis-classified by the previous trees. For RFs, the final classification output is typically obtained through a voting mechanism (e.g., majority voting), while GBDTs are usually trained for binary classification problems and use some binary reduction scheme, such as one-vs-all or one-vs-one, for multi-class classification.

Our \our{} trees are standard CARTs and we will use the random forest model for training an ensemble of \our{} trees. In our experimental evaluation, on the one hand we will compare \our{} models with natural RFs in order to show that \our{} training is able to make RFs robust, and, on the other hand, \our{} models will be also compared with the state-of-the-art of robust GBDTs, that is, the adversarially trained models by \cite{Andriushchenko19} and \cite{ChenZBH19}.

\section{Training as Combinatorial Optimization}
\label{sec:training}
Given a training dataset $D = \{(\vec{x}_1,y_1),\ldots, (\vec{x}_N,y_N)\} \subseteq  X\times\cL$, a training algorithm explores a hypothesis space $\mathcal{H}\subseteq X \rightarrow \wpp(\mathcal{L})$ searching for a classification model in $\mathcal{H}$ which maximizes some performance function $f: (X \rightarrow \wpp(\mathcal{L})) \rightarrow \mathbb{R}$. Training set, performance function and search strategy will determine how the output classification model is computed. In the case of CART decision trees, the training process first builds a tree stump consisting of a single internal node labeled by a univariate hard split $S_{i,k}\ud x_i \leq k$, where the attribute $i$ and the threshold $k$ are selected among all possible $i \in [1,d]$ and $k\in \bR$. Each split candidate $S_{i, k}$ yields two new leaves $\lambda_l$ and $\lambda_r$ which store a distribution frequency of training samples $(\#y_1, \ldots, \#y_m)\in \bN^m$ grouped by labels $y_i\in \cL$, i.e., equivalently, a probability distribution $(p_1, \ldots, p_m)\in \bR^m$ for labels which is used to compute either the entropy $h$ or Gini $g$ indexes which encode the information gain for a leaf $\lambda=(p_1, \ldots, p_m)$ as follows:
\begin{align*}
\textstyle
 h(\lambda) \ud  - \sum_{i = 1}^{m} p_i \cdot log_m(p_i)
 \qquad
 g(\lambda) \ud 1 - \sum_{i = 1}^{m} p_i^2
\end{align*}
These indexes are used to estimate the purity of a split $S_{i, k}$ by averaging their values on its  leaves $\lambda_l$ and $\lambda_r$ as follows:
\begin{align*}
 H(S_{i, k}) &\ud (|\lambda_l| h(\lambda_l) + |\lambda_r|h(\lambda_r))/(|\lambda_l| + |\lambda_r|)\\
 G(S_{i, k}) &\ud (|\lambda_l| g(\lambda_l) + |\lambda_r| g(\lambda_r))(|\lambda_l| + |\lambda_r|)
\end{align*}
where $|\lambda|$ denotes the number of samples reaching the leaf $\lambda$ for the split $S_{i, k}$. The training algorithm selects a split candidate $S_{i,k}$ which minimizes $H$ or $G$, and the process will be repeated until no more splits are possible, i.e.\ every leaf contains only samples with a same label, or other custom criteria are met such as maximum tree depth, maximum number of leaves or minimum number of samples per leaf (these are tunable parameters, e.g., with scikit-learn).

This training process therefore corresponds to a greedy search algorithm which tries to find a globally optimal tree by looking at local split information only. This approach exhibits two major drawbacks: 
\begin{enumerate}[{\rm (A)}]
 \item it does not take robustness (or stability) into account;
 \item it is limited by information from local splits only.
\end{enumerate}
These limitations may often lead to overfitting, which is a well-known phenomenon with decision trees \cite{Bramer2013,horning2013introduction}. Pruning techniques, e.g.\ \cite{bradford1998pruning,kearns1998fast,kijsirikul2001decision}, can be used to counteract overfitting by reducing the size of the tree in order to simplify the model, although this may often lead to a loss of accuracy and may be applied just as a post-training step.

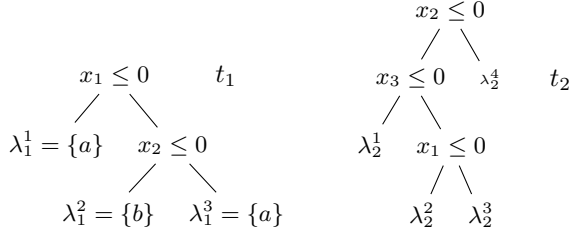
\begin{figure}[tb]
 \centering
 \small
 \begin{forest} 
  [$x_1 \leq 0$, name = n1
    [{$\lambda_1^1 =\{a\}$}]
    [$x_2 \leq 0$
      [{$\lambda_1^2=\{b\}$}]
      [{$\lambda_1^3=\{a\}$}]
    ]
  ]
  \node at (n1) [right=9ex]{{\normalsize $t_1$}};
 \end{forest}
 \qquad
 \begin{forest} 
  [$x_2 \leq 0$
    [$x_3 \leq 0$
      [{ $\lambda_2^1$}]
      [$x_1 \leq 0$
        [{ $\lambda_2^2$}]
        [{ $\lambda_2^3$}]
      ]
    ]
    [{\tiny $\lambda_2^4$}, name = n2]
  ]
    \node at (n2) [right=5ex]{{\normalsize $t_2$}};
 \end{forest}
 \caption{Two decision trees $t_1$ (left) and $t_2$ (right).}
 \label{fig:example-decision-tree}
\end{figure}

Of course, local properties of decision trees, in general, cannot be  lifted globally to the whole tree, as it is the case of stability in this simple example.
\begin{example}
 Consider the tree $t_1$ in Fig.~\ref{fig:example-decision-tree} and its right subtree $t_r$ rooted at the split node $x_2\leq 0$. Consider an input region $Y = \{(x_1, x_2) \in \mathbb{R}^2 ~|~ -1 \leq x_1 \leq 1,\, -1 \leq x_2 \leq 0\}$. Then, the set of reachable leaves from $Y$ in $t_r$ is labeled as $\{b\}$, hence ensuring stability of $t_r$ in $Y$, while the set of reachable leaves from $Y$ in  $t_1$ is labeled as $\{a, b\}$, thus making the classification of the tree $t_1$ not stable in $Y$.  
 \qed
\end{example}

We will design a global tree training algorithm which takes stability into account. Given a (finite) set $M$ of performance metrics of type $\cH \rightarrow \mathbb{R}$, such as accuracy on a given test set and stability on a given test set w.r.t.\ a perturbation,
we combine them through a comprehensive performance function $\varphi_M: \cH \rightarrow \mathbb{R}$ that will be maximized during the training process. For example, this performance function  can be a simple linear combination $\varphi_M(C) = \sum_{i=1}^{|M|} w_im_i(C)$. 
If $\varphi_M$ is differentiable and, for some  metric $m_k \in M$, 
$\frac{\partial \varphi_M}{\partial m_k} \geq 0$, then  a classifier $C_{\mathrm{opt}}$ which is computed by an ideal trainer by maximizing $\varphi_M$ will be optimal for the performance metric $m_k$, meaning  that for any classifier $C\in \cH$, if $m_j(C_{\mathrm{opt}})=m_j(C)$ for all $j\neq k$, then $m_k(C_{\mathrm{opt}}) \geq m_k(C)$. 

Of course, metrics such as accuracy and stability can only be estimated on a finite test subset of the input space $X$, meaning that global optima cannot be computed precisely, so that training algorithms effectively will find classifiers which are locally optimal for a finite test subset $T$ of $X$. Finding a locally optimal classifier for $T$ is not necessary in practice, as local optimality is unlikely to be an appropriate measure of global optimality and could even lead to overfitting phenomena. We argue that heuristic search strategies such as genetic algorithms may provide a viable and effective training procedure of tree classifiers which closely approximate the ideal optimal solutions and may improve the state-of-the-art in practice.

\section{Genetic Adversarial Training by Meta-Silvae}
\label{sec:genetic-algorithm}

A significant number of training procedures for decision trees based on evolutionary algorithms have been investigated, as surveyed by \cite{BarrosBCF12}. The underlying basic idea is that evolutionary algorithms perform a robust global, as opposed to local, search in the hypothesis space and tend to cope better with relationships between different features  than greedy methods~\cite{freitas}. To the best of our knowledge, we put forward the first adversarial training procedure for decision tree (or tree ensemble) classifiers based on a genetic algorithm, called Meta-Silvae (\our{}), which targets to maximize both accuracy and stability metrics. \our{} crucially relies on a complete  formal verification method of the stability (or robustness) of ensembles of decision trees.

\subsection{Complete Verification of Stability}
We design and implement  a decision tree adversarial training method based on a genetic algorithm, that we call \our{} (for Meta-Silvae). This genetic learning algorithm internally relies on silva~\cite{RZ20}, an open source tool based on abstract interpretation \cite{CC77,rival-yi} for the formal verification of stability properties of decision tree ensembles. Silva performs a static analysis of ensembles of decision trees in an \emph{abstract domain} $A\subseteq \wp(\bR^d)$ which represents some selected properties of real vectors, such as hyperrectangles of intervals providing lower and upper bounds to vector components or 
a domain of linear relations between vector components. Silva first approximates in the abstract domain $A$ an adversarial region $P(\vec{x})\in \wp(\bR^d)$ of an input vector $\vec{x}\in \bR^d$ and then abstractly computes a sound \emph{overapproximation} of  the set of leaves of a decision tree (or an ensemble of trees) which are reachable from the adversarial samples ranging in $P(\vec{x})$. This static analysis is based on the \emph{soundness} principle of abstract interpretation meaning that no leaf reachable from some sample in $P(\vec{x})$ can be missed. When adversarial attacks are modeled by the maximum norm perturbation $P_{\infty, \epsilon}(\vec{x})$ and the analysis is performed on the abstract domain of hyperrectangles the output returned by silva turns out to be \emph{complete}, meaning that each leaf computed by silva is actually reached by some adversarial input ranging in $P_{\infty, \epsilon}(\vec{x})$, namely, no false positive (i.e., a false reachable leaf) may happen. Silva therefore provides a complete certification algorithm for the stability (or robustness) of an input sample $\vec{x}$ under adversarial attacks in $P_{\infty, \epsilon}(\vec{x})$ and, in turn, this verification tool allows us to derive \emph{precisely} (thanks to completeness) the stability $\mathit{st}_{T, P_{\infty, \epsilon}}$ and robustness $\mathit{rob}_{T, P_{\infty, \epsilon}}$  metrics, as defined in Section~\ref{sec:background}, for a tree ensemble classifier 
on some test set $T$.

To conclude this brief outline of silva, let us recall that the abstract domain of hyperrectangles $\HR_d$ consists of $d$-dimensional vectors of intervals $[l,u]$, where the bounds $l,u\in \mathbb{R}\cup \{+\infty, -\infty\}$ are such that $l\leq u$, for example 
$([0.1,0.9], [-\infty,0],[0.5,+\infty])\in \HR_3$. Therefore, a hyperrectangle $[\vec{l},\vec{u}]=([\vec{l}_1,\vec{u}_1],...,[\vec{l}_d,\vec{u}_d])\in \HR_d$ represents all the real vectors $\vec{x} \in \mathbb{R}^d$ such that, for all $j\in [1,d]$, $\vec{l}_j\leq \vec{x}_j\leq \vec{u}_j$, i.e., such that lower/upper bounds of their components are correctly approximated by $[\vec{l},\vec{u}]$.

\subsection{\our{} Algorithm}\label{subsection:GAT}
By exploiting the  stability information provided by silva, for training on a dataset $T$ 
a decision tree in a hypothesis
space $\cH$,
we consider as objective performance function 
$\varphi_T: \cH \rightarrow \mathbb{R}$ a linear combination of accuracy and stability, i.e.,
$\varphi_T(t) \ud w_a \textit{acc}_T(t) + w_s \textit{st}_{T,P}(t)$ for a given perturbation $P$ and for some weights $w_a,w_s\in \bR$. Since accuracy and stability are independent of each other, for nonnegative weights $w_a, w_s\geq 0$, $\varphi_T$ is differentiable and such that $\frac{\partial \varphi_T}{\partial \textit{acc}_T} =w_a\geq 0, \frac{\partial \varphi_T}{\partial \textit{st}_{T,P}}=w_s \geq 0$, so that an ideal training algorithm which maximizes $\varphi_T$ will output optimal decision trees. For instance, in the training example depicted in Fig.~\ref{fig:train-standard-vs-our} we used as objective function $\varphi(t) \ud 0.9 \textit{acc}(t) + 0.1 \textit{st}_{\pm2\%}(t)$.

\begin{algorithm}[htb]
 \textit{population} = generateInitialTrees()\;
 \textit{nextPopulation} = $\varnothing$\;
 \While{$\neg$\,{\rm{stopCriterion}()}}{
 \tcp*[h]{{\sc Elitist Selection}}
 
  \textit{nextPopulation}.push(\textit{population}.selectBest($\varphi_T$))\;
  \While{$\neg$\,\textit{nextPopulation}.{\rm isFull()}}{
   \tcp*[h]{{\sc Roulette Wheel Selection}}
   
   \textit{parentA} = \textit{population}.select($\varphi_T$)\;
   \textit{parentB} = \textit{population}.select($\varphi_T$)\;
   \tcp*[h]{{\sc Single-Point Crossover}}
   
   \textit{offspring} = crossover(\textit{parentA}, \textit{parentB})\;
   \tcp*[h]{{\sc Mutation With Probability }$p$}
   
   \If{\textit{offspring}.{\rm shouldMutate($p$)}}{
     \tcp*[h]{{\sc Grow or Grow-Or-Prune}}
   
    \textit{offspring}.mutate()\;
   }
   \textit{nextPopulation}.push(\textit{offspring})\;
  }
  \textit{population} = \textit{nextPopulation}\;
  \textit{nextPopulation} = $\varnothing$
 }
 \Return \textit{population}.selectBest($\varphi_T$)
 \caption{\textbf{\our{}}}
 \label{alg:our}
\end{algorithm}

Our suboptimal solution is computed by the genetic Algorithm~\ref{alg:our} called \our{}. At line~1, a set of trees is generated and stored as initial \textit{population}, where \our{} can either start with a set of base trees consisting of a single leaf only (which is our choice in the experimental evaluation) or by any set of pre-trained decision trees. Then, the while-loop at lines 3-13 generates the \textit{nextPopulation} and iterates until some stop criterion, such as a timeout or a bound on the \textit{population} size, is met. At each iteration, the best tree w.r.t.\ our performance function $\varphi_T$ will be first selected at line~4 from the current population to carry over to the next one unchanged. This therefore implements an elitist selection strategy for GAs \cite{Baluja} which ensures that the performance function will not decrease from the current population to the next. The new individuals will be generated by the while-loop at lines~5-11, where two trees \textit{parentA} and \textit{parentB} are first selected from the current population, then combined in an offspring by a crossover operation, which can finally be mutated for enhancing genetic diversity. 

\paragraph{\textbf{\textit{Selection.}}}
A fitness proportional selection scheme \cite{neumann} is employed at lines 6-7, also known as roulette wheel, which selects an individual $t_i$ from a current population $Y$ with a probability $p_i$ which is proportional to its fitness as determined by the objective function $\varphi_T$ as follows: $p_i \ud \frac{\varphi_T(t_i)}{\sum_{t \in Y}\varphi_T(t)}$. It should be remarked that in this selection \our{} exploits crucially the stability value  $\textit{st}_{T,P}(t)$ given by the verification tool silva which is used by $\varphi_T(t)$. 

\paragraph{\textbf{\textit{Crossover.}}}
The information of the two selected trees \textit{parentA} and \textit{parentB} is combined at line~8 by the following variation of the standard single-point crossover procedure \cite{crossover}:
\begin{enumerate}[{\rm 1.}]
 \item a subtree $t_A$ is randomly selected and pruned from \textit{parentA};
 \item a subtree $t_B$ is randomly selected from \textit{parentB};
 \item $t_B$ is inserted as subtree of \textit{parentA} in place of the pruned subtree $t_A$; 
 \item consistency of the new tree is enforced by pruning those nodes which do not contain information and then by reshaping the tree as needed.
\end{enumerate}
Let us illustrate this crossover function by a simple example. 

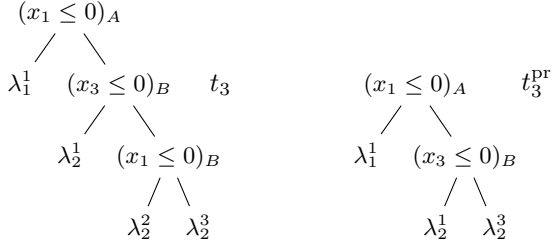
\begin{figure}[t]
 \centering
 \small
 \begin{forest} 
  [$(x_1 \leq 0)_A$
    [$\lambda_1^1$]
    [$(x_3 \leq 0)_B$, name=n1
      [$\lambda_2^1$]
      [$(x_1 \leq 0)_B$
        [$\lambda_2^2$]
        [$\lambda_2^3$]
      ]
    ]
  ]
  \node at (n1) [right=8ex]{{\normalsize $t_3$}};
 \end{forest}
 \qquad\qquad
 \begin{forest} 
  [$(x_1 \leq 0)_A$, name = n2
    [$\lambda_1^1$]
    [$(x_3 \leq 0)_B$
      [$\lambda_2^1$]
      [$\lambda_2^3$]
    ]
  ]
    \node at (n1) [right=7ex,above=5ex]{{\normalsize $t_3^{\mathrm{pr}}$}};
 \end{forest}
 \caption{A decision tree $t_3$ (left) and its pruning $t_3^{\mathrm{pr}}$ (right).}
 \label{fig:example-decision-tree-2}
\end{figure}

\begin{example}
 Consider as parent trees $t_1$ and $t_2$ in Fig.~\ref{fig:example-decision-tree}, and assume that the nodes labeled with splits $x_2 \leq 0$ and $x_3 \leq 0$ are selected, resp., as subtrees $t_A$ of $t_1$ and $t_B$ of $t_2$. The tree $t_B$ is inserted in $t_1$ in place of $t_A$, as depicted by the left tree $t_3$ in Fig.~\ref{fig:example-decision-tree-2}. However, after this insertion, the path $(x_1 \leq 0)_B \ra \lambda_2^2$ becomes unfeasible due to the parent constraint $(x_1 > 0)_A$ at the root, so that the subtree of $t_3$ rooted at $(x_1 \leq 0)_B$  must be pruned of the left path $(x_1 \leq 0)_B \ra \lambda_2^2$, thus yielding the output tree $t_3^{\mathrm{pr}}$ on the right of Fig.~\ref{fig:example-decision-tree-2}. \qed
\end{example}

\paragraph{\textbf{\textit{Mutation.}}}
After crossover, mutation of the offspring happens at lines~9-10 with probability $p$ and can either prune a randomly selected subtree or transform a leaf into a single split. In the latter case, consistency of splits must be preserved so that logically inconsistent trees are never generated. The mutation probability $p$ is specified as a tunable parameter of \our{}, as well as the adopted mutation strategy which can be chosen between \emph{grow} only and \emph{grow-or-prune}. In both cases the selection is made on a stochastic basis by considering the entropy $h$ of each node (or, alternatively, the Gini index $g$). We describe this mutation procedure more in detail through an example.

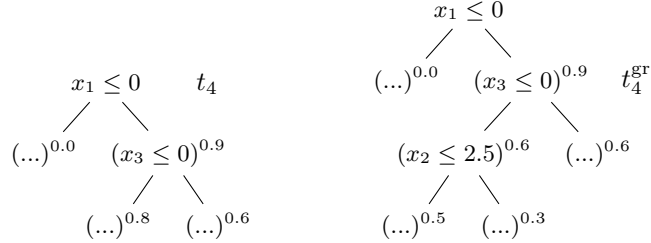
\begin{figure}[t]
 \small
\centering
 \begin{forest}
  [$x_1 \leq 0$, name=n1
    [${(...)^{0.0}}$]
    [${(x_3 \leq 0)}^{0.9}$
      [${(...)^{0.8}}$]
      [${(...)^{0.6}}$]
    ]
  ]
  \node at (n1) [right=8ex]{{\normalsize $t_4$}};
 \end{forest}
 \qquad\qquad
 \begin{forest}
  [$x_1 \leq 0$ 
    [${(...)^{0.0}}$]
    [${(x_3 \leq 0)}^{0.9}$, name =n4
      [${(x_2 \leq 2.5)}^{0.6}$
        [${(...)^{0.5}}$]
        [${(...)^{0.3}}$]
      ]
      [${(...)^{0.6}}$]
    ]
  ]
   \node at (n4) [right=8ex]{{\normalsize $t_4^{\mathrm{gr}}$}};
 \end{forest}
\caption{An example of tree mutation.}\label{examplemutation}
\end{figure}

\begin{example}
Consider the decision tree $t_4$ on the left of Fig.~\ref{examplemutation} whose nodes are decorated as superscripts with their entropy values. 
A grow mutation starts from the root node $x_1\leq 0$, computes the entropy $h$ of its left and right children, and move toward one of them with a probability which is proportional to their respective entropies, in our example the node $(x_3\leq 0)^{0.9}$ is chosen. Once a leaf is reached, in our example $(...)^{0.8}$, the samples associated to that leaf are used to generate a set of features $i$ and thresholds $k$ which defines a set of split candidates $x_i \leq k$. 
The split candidate maximizing the objective function $\varphi_T$ is then 
selected by applying the split and evaluating accuracy and stability of the corresponding tree. For example, the split $x_2 \leq 2.5$ is selected for growing the leaf $(...)^{0.8}$ 
and this yields 
the tree $t_4^{\mathrm{gr}}$ depicted on the right of Fig.~\ref{examplemutation}.
\qed 
\end{example}

It should be remarked that while the standard learning method for CART trees
is a greedy algorithm which
incrementally builds a decision tree by locally computing new split nodes or new leaves~\cite{BreimanFOS84},
in \our{} the selection of a candidate split %
relies on a stability test (performed by silva) on the whole corresponding candidate tree, thus meaning that the
\our{} learning
process is inherently not incremental and consequently could be computationally burdensome.
We therefore introduced in \our{} an  optimization parameter, called ``aggressiveness''
in the implementation of \our{},
which allows us to set up a threshold on the number of split candidates which are immediately evaluated during the mutation phase, while those exceeding this threshold are delayed. 
It turns that this optimization is effective in reducing the computational burden 
without sacrificing the overall generalization, because the evaluation of split candidates which are ruled out by this threshold is delayed to later iterations of \our{}. 

For the \emph{grow-or-prune} mutation, a similar strategy is applied. Starting from the root node, the mutation algorithm iteratively moves either to the right of left children with a probability which is proportional to their entropy $h$ and, at any point, the current node has a probability of $1 - h$ of being pruned. By doing so, splits with low entropies are likely to be pruned, thus improving the tree compactness and stability at the bearable cost of losing some accuracy. 
If no pruning happens then the mutation proceeds as in the \emph{grow} case.

\paragraph{\textbf{\textit{Output.}}}
At the end of the whole evolutionary process, the best tree classifier of the final population, which is also the best tree encountered during the whole process due to elitism, is returned as output decision tree.
As an example, the diagrams in Fig.~\ref{fig:train-standard-vs-our} have been generated by the following two decision trees, where the first one has been trained by the standard algorithm of scikit-learn and the second one by \our{}. In both trees, a leaf $\lambda = {(n{\color{red}^\bullet}, m{\color{blue}^{{\textbf{+}}}})}$ denotes that $\lambda$ stores 
$n$ red bullet samples and $m$ blue cross samples.

{\center
 \resizebox{0.8\columnwidth}{!}{
 \begin{forest}
[$x_1 \leq 0.555$
  [$x_2 \leq 0.665$, name=n2
    [$x_2 \leq 0.555$
      [${(0{\color{red}^\bullet}, 25{\color{blue}^{{\textbf{+}}}})}$]
      [$x_1 \leq 0.335$
        [${(0{\color{red}^\bullet}, 5{\color{blue}^{{\textbf{+}}}})}$]
        [${(3{\color{red}^\bullet}, 0{\color{blue}^{{\textbf{+}}}})}$]
      ]
    ]
    [${(22{\color{red}^\bullet}, 0{\color{blue}^{{\textbf{+}}}})}$]
  ]
  [$x_2 \leq 0.105$
    [$x_2 \leq 0.050$
      [${(1{\color{red}^\bullet}, 0{\color{blue}^{{\textbf{+}}}})}$]
      [${(0{\color{red}^\bullet}, 1{\color{blue}^{{\textbf{+}}}})}$]
    ]
    [$x_1 \leq 0.625$
      [$x_1 \leq 0.605$
        [${(4{\color{red}^\bullet}, 0{\color{blue}^{{\textbf{+}}}})}$]
        [${(0{\color{red}^\bullet}, 1{\color{blue}^{{\textbf{+}}}})}$]
      ]
      [${(38{\color{red}^\bullet}, 0{\color{blue}^{{\textbf{+}}}})}$]
    ]
  ]
]
\node at (n2) [left=5ex,above=4ex]{\textbf{scikit-learn}};
 \end{forest}
}
 \\%
 \qquad
 { %
 \resizebox{0.8\columnwidth}{!}{
  { %
 \quad\begin{forest} 
[$x_2 \leq 0.412$
  [$x_1 \leq 0.685$, , name=n2
    [${(0{\color{red}^\bullet}, 23{\color{blue}^{{\textbf{+}}}})}$]
    [$x_1 \leq 0.765$
      [${(2{\color{red}^\bullet}, 0{\color{blue}^{{\textbf{+}}}})}$]
      [${(10{\color{red}^\bullet}, 0{\color{blue}^{{\textbf{+}}}})}$]
    ]
  ]
  [$x_1 \leq 0.478$
    [$x_2 \leq 0.664$
      [$x_1 \leq 0.365$
        [${(0{\color{red}^\bullet}, 7{\color{blue}^{{\textbf{+}}}})}$]
        [$x_2 \leq 0.525$
          [${(0{\color{red}^\bullet}, 2{\color{blue}^{{\textbf{+}}}})}$]
          [${(1{\color{red}^\bullet}, 0{\color{blue}^{{\textbf{+}}}})}$]
        ]
      ]
      [${(19{\color{red}^\bullet}, 0{\color{blue}^{{\textbf{+}}}})}$]
    ]
    [${(36{\color{red}^\bullet}, 0{\color{blue}^{{\textbf{+}}}})}$]
  ]
]
\node at (n2) [left=5ex,above=4ex]{\textbf{\our{}}};
  \end{forest}
  }
  }
  }
}

\subsection{Ensembles of \our{} Trees}
\label{sec:forest-training}
Multiple instances of \our{} can be run (also in parallel: this chance was exploited in our experimental evaluation on a cluster) in order to generate a tree ensemble, which in our experiments is a random forest. When training a random forest, we need to achieve a significant random diversification among its trees, typically obtained by a random sampling of the set of features which each tree can explore when searching for a new split. \our{} allows to specify the number $N$ of features which can be inspected for training a tree. A subset of features of size $\leq N \in [1,d]$ is randomly extracted, where each feature has the same probability of being selected. This feature selection happens only once before the training, and it is therefore applied tree-wise by \our{}. The resulting set of trees can be used as a single classifier by applying standard voting mechanisms (such as a simple majority vote) used for natural random forests.

\section{Experimental Evaluation}
\label{sec:experimental-evaluation}
We implemented \our{} as a self-contained C program whose source code (about 3K LOC) together with datasets, classification models and scripts is publicly 
available on GitHub \cite{metasilvae}. Our experiments were run on a cluster of 15 computing nodes, each equipped with two Intel Xeon CPU E5520 at 2.27GHz with 8 cores and 32GB RAM. In our experiments we considered the collection of standard datasets with numerical features summarized in Table~\ref{tab:datasets}: these datasets are used in~\cite{Andriushchenko19,ChenZBH19,TornblomN20}, while wine is a UCI dataset~\cite{uci}. 

\begin{table}[htb]
 \centering
 \resizebox{0.9\columnwidth}{!}{
 \begin{tabular}{| l | c c c c c |}
  \hline
  \textbf{Dataset} & \#\textbf{classes} & \#\textbf{features} & \textbf{values}& \#\textbf{train} & \#\textbf{test}\\
  \hline
  breast-cancer                 &  2 &  10 &  [1,10] &   546 & 137\\
  cod-rna                       &  2 &   8 &   [0,1] & 59535 & 271617\\
  collision-detection           &  2 &   6 &   [0,1] & 30000 & 3000\\
  diabetes                      &  2 &   8 &   [0,1] &   614 & 154\\
  fashion-mnist                 & 10 & 784 & [0,255] & 60000 & 10000\\
  ionosphere                    &  2 &  34 &  [-1,1] &   260 & 90\\
  mnist                         & 10 & 784 & [0,255] & 60000 & 10000\\
  mnist-1-5                     &  2 & 784 & [0,255] & 12162 & 2026\\
  mnist-2-6                     &  2 & 784 & [0,255] & 11875 & 1989\\
  sensorless                    & 11 &  48 &   [0,1] & 48509 & 10000\\
  wine                          &  2 &  13 &   [0,1] &   128 & 50\\
  \hline
 \end{tabular}
 }
 \caption{Summary of Datasets.}
 \label{tab:datasets}
\end{table}

\subsection{Setup}
\begin{figure}[htb]
 \centering
 \includegraphics[width=0.8\textwidth]{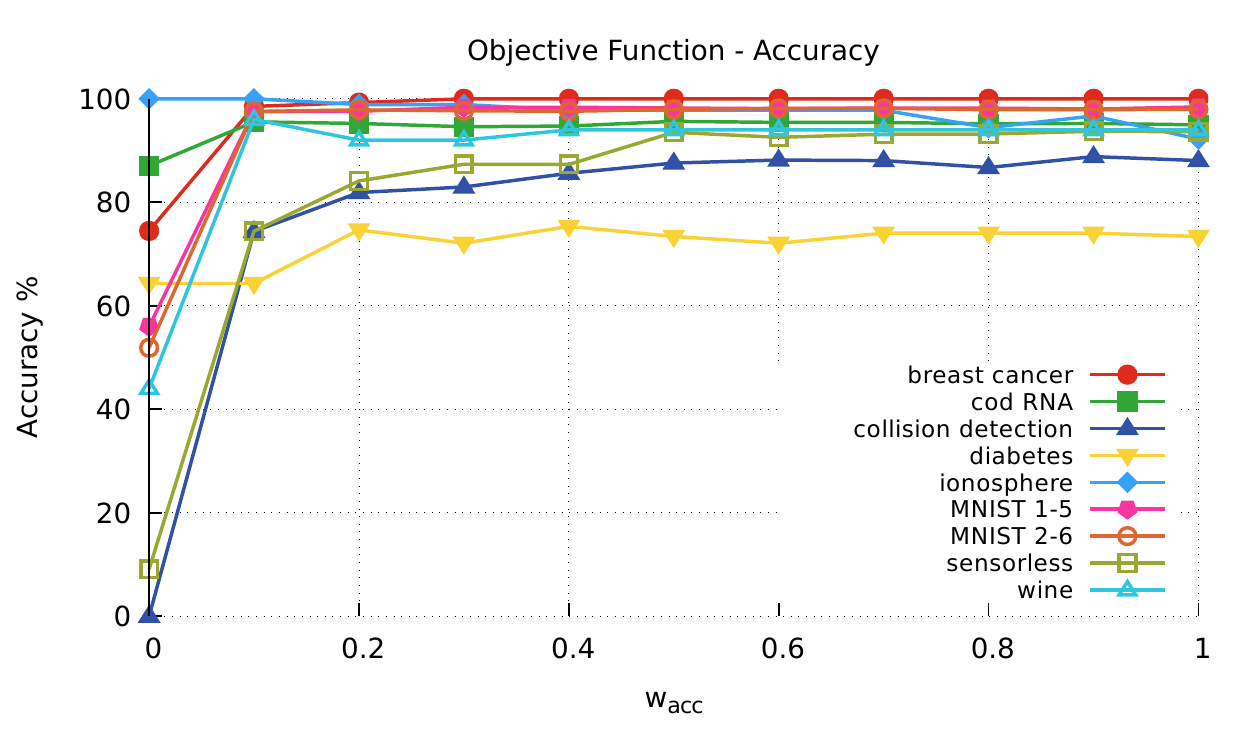}
 
 \includegraphics[width=0.8\textwidth]{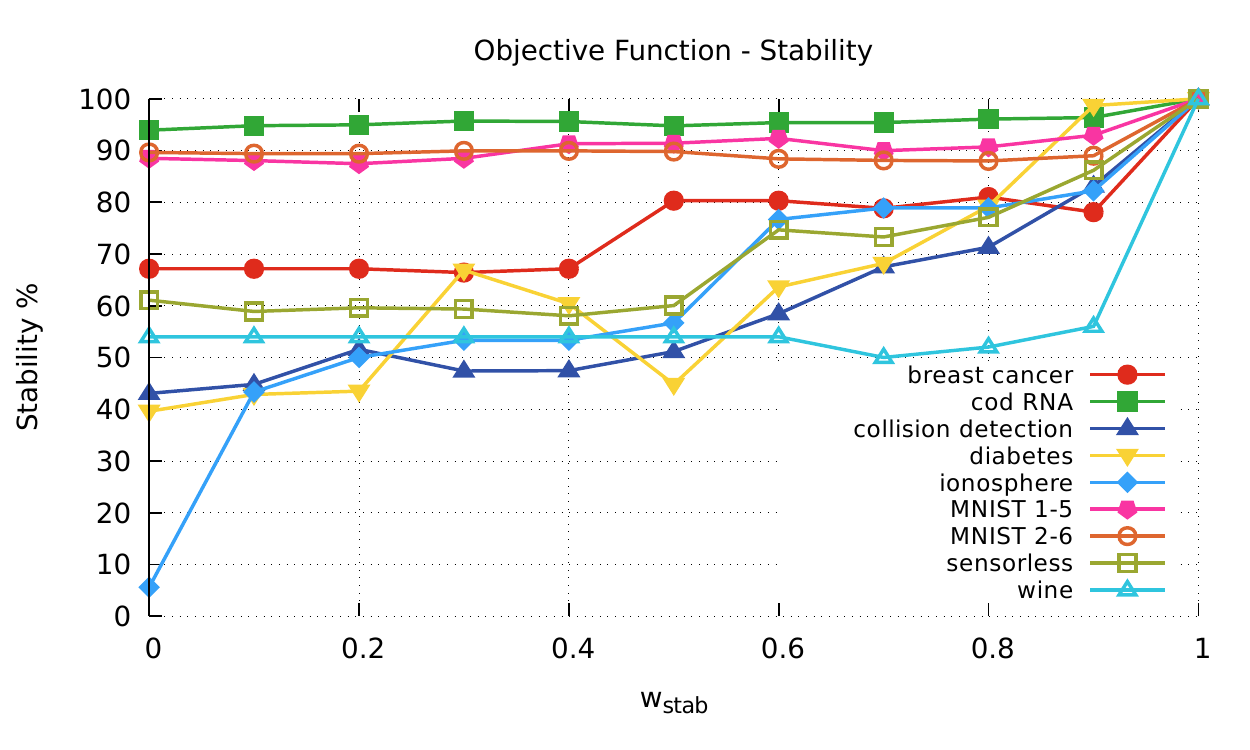}
 \caption{Impact of Different Weights.}
 \label{fig:objective-function}
\end{figure}

\our{} does not require a specific 
tuning of hyperparameters for training decision trees, although the choice of the objective function $\varphi$ plays a fundamental role. As described in Section~\ref{subsection:GAT},  we adopted a weighted linear combination of accuracy and stability given by 
$\varphi_T(t) = w_{\text{acc}} \textit{acc}_T(t) + w_{\text{stab}} \textit{st}_{T,P}(t)$, for a given 
training dataset $T$, 
perturbation $P$ and weights $w_{\text{acc}},w_{\text{stab}}\in [0,1]$ such that $w_{\text{acc}}+w_{\text{stab}} = 1$. 
Fig.~\ref{fig:objective-function} shows the impact of selecting different weights (where $w=0.1k$ for $k\in \{0,...,10\}$)
on the final tree classifier in terms of accuracy and stability on the whole training set $T$. 

As expected, higher values of $w_{\text{acc}}$ tend to yield more accurate models, where we observed a converging behaviour when accuracy exceeds $70\%$. On the other hand, stability increases with higher $w_{\text{stab}}$  weights, it is above $45\%$ and often above
$60\%$ already with  $w_{\text{stab}}=0.1$,
 although stability generally may display a more varying behaviour than accuracy. These experiments hint that an effective and well balanced choice of weights can be obtained with
$w_{\text{acc}} = 0.9, w_{\text{stab}} = 0.1$. It is worth remarking that by setting $w_{\text{stab}} = 1$ (and therefore $w_{\text{acc}} = 0$), as expected, \our{} will always produce perfectly stable models corresponding to a constant function, which is generally (but not always) inaccurate. 

\begin{figure}[htb]
 \centering
 \includegraphics[width=0.85\textwidth]{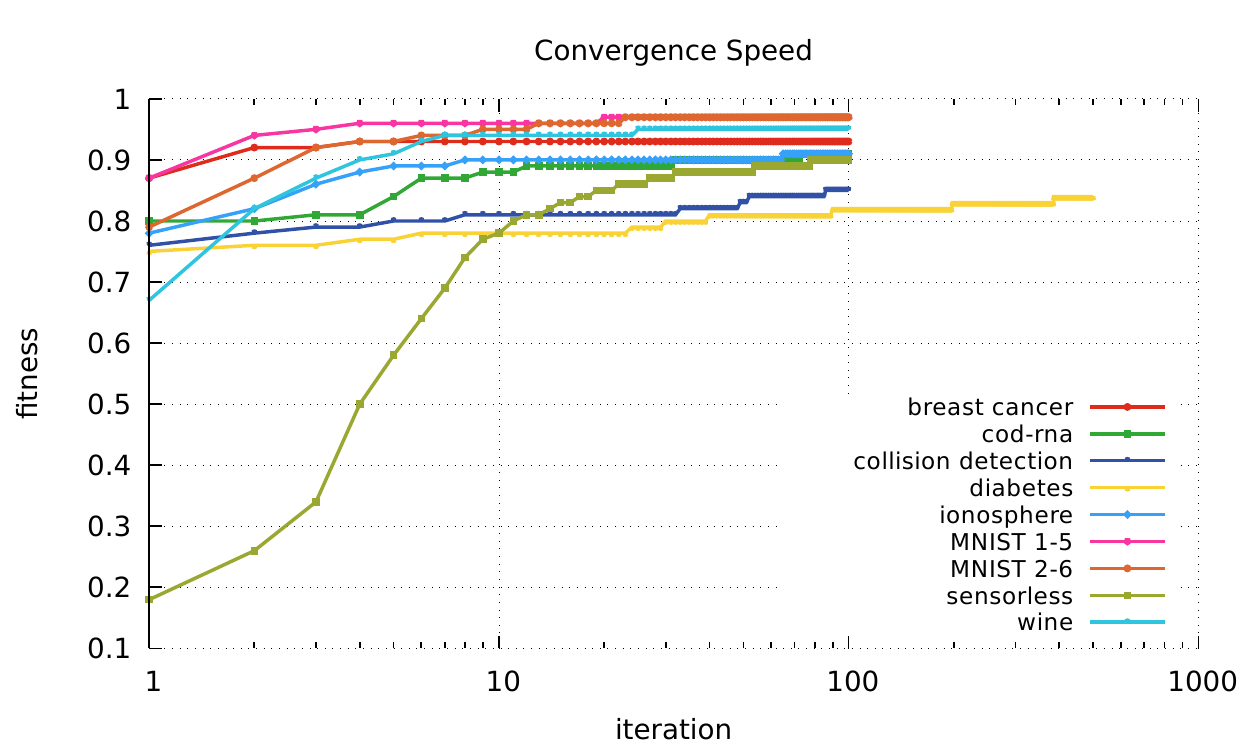}
 \caption{Convergence Speed.}
 \label{fig:convergence}
\end{figure}

We analysed the rate of convergence of the fitness function $\varphi_T$ along iterations of \our{} training. 
Fig~\ref{fig:convergence} displays the fitness of the best individual of $\textit{population}$ 
after each iteration of \our{}, where the number of iterations is shown on a logarithmic scale. We observe that the best gains in fitness are achieved during the first iterations and \our{} tends to converge within 50-70 iterations, with the exception of the dataset diabetes
which needs about 500 iterations to converge. Hence, the maximum number of iterations has been set to $500$ for diabetes and to $100$ for the other datasets. This same method is used to decide between \emph{grow} and \emph{grow-or-prune} mutations as well as their 
``aggressiveness'' threshold, as described in Section~\ref{subsection:GAT}.

\subsection{Results}
Table~\ref{tab:training} summarizes the main  data of our \our{} models together with their accuracy and stability metrics. \ours{} have been trained for the objective performance function $\varphi_T(t) = 0.9 \textit{acc}_T(t) + 0.1 \textit{st}_{T,P_{\infty, \epsilon}}(t)$ where stability is 
w.r.t.\ a maximum norm perturbation $P_{\infty, \epsilon}$ whose magnitudes $\epsilon$ are displayed as absolute and percentage values. These magnitudes $\epsilon$ are the same used by \cite{Andriushchenko19,ChenZBH19} for their adversarial training on 6 common datasets: breast-cancer, cod-rna, diabetes, fashion-mnist, mnist-1-5, mnist-2-6. It is worth remarking that some of these perturbations are quite challenging, since they peak to $\pm 30.2\%$ for \mbox{mnist-1-5/2-6} and $\pm 33.3\%$ for breast-cancer. For fashion-mnist and mnist we trained a random forest of 50 \our-robust trees adopting a standard majority voting for output classification. For the remaining datasets it was enough to train a single robust decision tree since in these cases random forests do not bring practical benefits in accuracy or stability, thus making these single tree models efficient and easily interpretable. It is  worth observing that \our{} training times are modest and acceptable even for the more demanding datasets fashion-mnist and mnist. The stability metrics w.r.t.\ $P_{\infty, \epsilon}$ have all been computed with no imprecision by silva on the whole test sets.

\begin{table}[htb]
\centering
 \resizebox{0.9\columnwidth}{!}{
 \begin{tabular}{| l | c c c c | c c|}
 \hline
 \multirow{3}{*}{\textbf{Dataset}}  & \multicolumn{4}{c|}{\textbf{\our{} training}}&\multicolumn{2}{c|}{\textbf{\our{} metrics}}\\
 \cline{2-7}
 & {\text{\#~of}} & {\text{max}} & \multirow{2}{*}{$P_{\infty, \epsilon}$ ($\%$)} &{\text{time(s)}}
 & \multirow{2}{*}{\textit{acc}\%} & \multirow{2}{*}{\textit{st}\%}
 \\
 & {\text{trees}} & {\text{depth}} &  &{\text{per tree}} &  & 
 \\
 \hline
 breast-cancer   &  1 &  6 &    3 ($\pm 33.3\%$) &    0.2 & 100.0 & 89.1\\
 cod-rna         &  1 & 12 & 0.025 ($\pm 2.5\%$) &   20.0 &  95.6 & 89.9\\
 collision-det.\ &  1 & 12 &    0.1 ($\pm 10\%$) &   28.8 &  87.5 & 45.4\\
 diabetes        &  1 & 15 &    0.05 ($\pm 5\%$) &    2.1 &  76.0 & 68.8\\
 fashion-mnist   & 50 & 22 &     25 ($\pm 10\%$) & 3593.1 &  86.4 & 46.9\\
 ionosphere      &  1 & 10 &    0.2 ($\pm 10\%$) &    0.2 &  97.8 & 84.4\\
 mnist           & 50 & 20 &   30 ($\pm 11.8\%$) & 3376.4 &  95.6 & 81.7\\
 mnist-1-5       &  1 & 11 &   77 ($\pm 30.2\%$) &   14.4 &  97.9 & 94.0\\
 mnist-2-6       &  1 & 10 &   77 ($\pm 30.2\%$) &   11.7 &  98.1 & 88.7\\
 sensorless      &  1 & 15 &    0.01 ($\pm 1\%$) &  133.0 &  94.7 & 57.4\\
 wine            &  1 &  5 &    0.1 ($\pm 10\%$) &    0.1 &  92.0 & 68.0\\
 \hline
 \end{tabular}
 }
 \caption{Performance of \our}
 \label{tab:training}
\end{table}

\begin{figure}[t]
 {\centering
 \hspace*{-3ex}\includegraphics[width=0.55\linewidth]{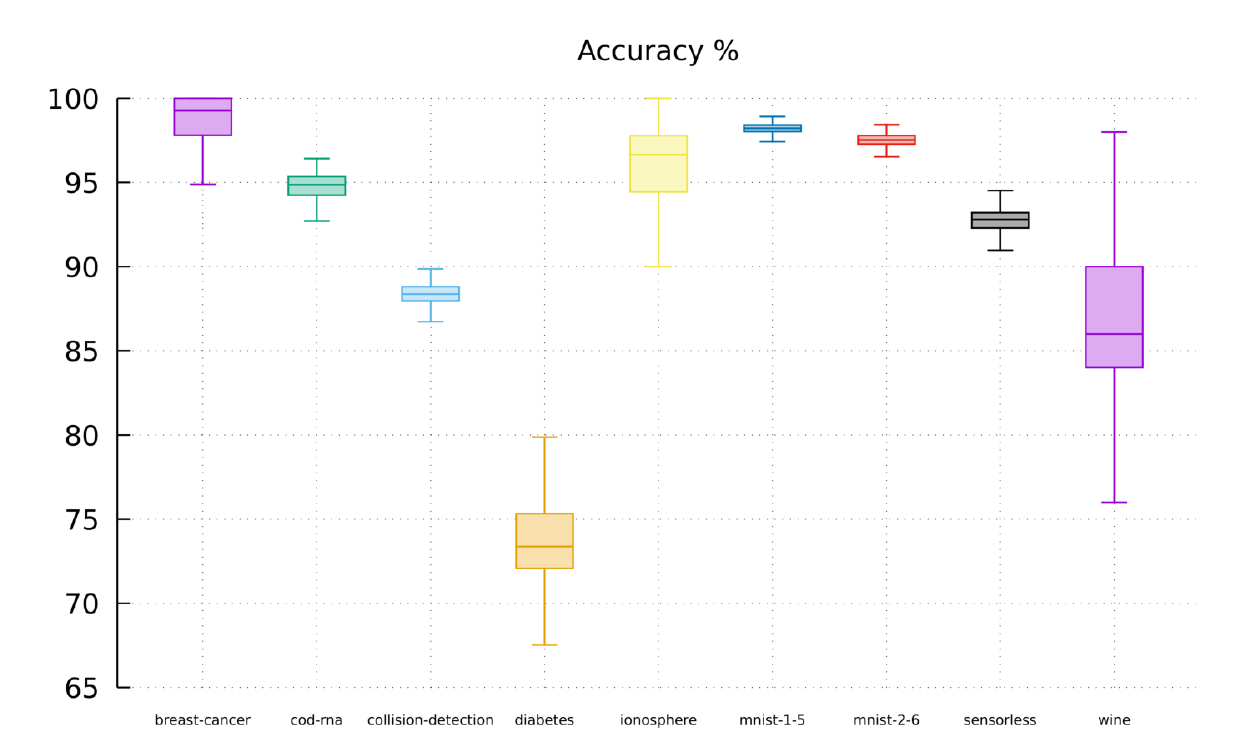}
 \hspace*{-2ex}
 \includegraphics[width=0.55\linewidth]{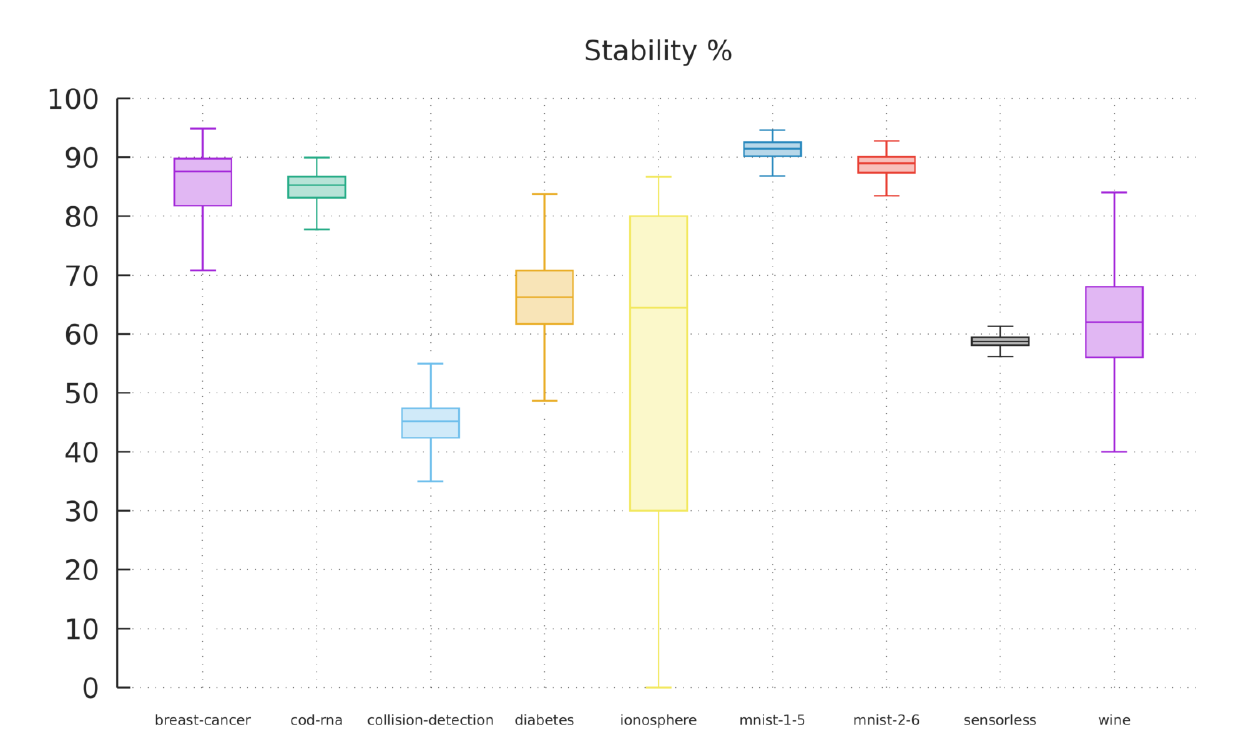}
 }
 \caption{Distribution of Accuracy/Stability over 1000 Runs.}
 \label{fig:variance}
\end{figure}

Since \our{} utilizes  a seed provided  by a random number generator, in order to analyse
the impact of randomness on the output models, the \our{} training has been repeated
1000 times for each dataset by selecting distinct values of the seed. The distribution 
of accuracy and stability of the models generated by these 1000 different runs is depicted by two box plot diagrams in Fig.~\ref{fig:variance}. The box is drawn from first to third quartiles, its horizontal line denotes the median. 
We observe that accuracy of most models is within $\pm 5\%$ (and often $\pm 2.5\%$) 
of the median, with the sole exception 
of wine.  
The results for stability are similar, where the stability of the models 
turns out to be within $\pm 5\%$ of the median, with the exception of ionosphere, where unstable (meaning stability $<30\%$) models can be produced, although 75\% of these 1000 trees has a stability $\geq 30\%$, and $\geq 65\%$ for half of them.

\subsection{\our{} vs RF}
We compared our robust models trained by \our{} with natural random forests trained by scikit-learn \cite{scikit-learn}. Hyperparameters for training random forests have been selected by a randomized grid search on the number of trees ranging in the 
interval $[5, 100]$, maximum depth in $[5, 100]$, and either Gini $G$ or entropy $H$ for split purity. By relying on the stability verification by 
silva~\cite{RZ20}, we selected the hyperparameters which maximize the same
objective function $\varphi_T$ used in \our{} training. 
The first table in Figure~\ref{tab:vs-rf} shows the accuracy and stability metrics on the whole test sets for RF as compared to \our{} models, where, for the sake of fair comparison, in the average of the relative stability gains of \our{} models we excluded the RFs with 0\% stability.

\begin{figure}[t]
\centering
\resizebox{0.9\columnwidth}{!}{
\begin{tabular}{| l | r r | r r | r r |}
\hline
\multirow{2}{*}{\textbf{Dataset}} & \multicolumn{2}{c |}{\textbf{RF}} & \multicolumn{2}{c |}{\textbf{\our{}}} & \multicolumn{2}{c |}{\textbf{Stability Gain}} \\
   & \textit{acc}\% & \textit{st}\% & \textit{acc}\% & \textit{st}\% &  abs.\ & rel.\  \\
\hline
breast-cancer   & \textbf{100.0} & 10.2 & \textbf{100.0} & \textbf{89.1} & +78.9\% & 8.7$\times$ \\
cod-rna         &  \textbf{97.9} & 62.3 &           95.6 & \textbf{89.9} & +27.6\% & 1.4$\times$ \\
collision-det.\ &  \textbf{94.8} & 21.0 &           87.5 & \textbf{45.4} & +24.4\% & 2.2$\times$ \\
diabetes        &  \textbf{78.6} & 20.8 &           76.0 & \textbf{68.8} & +48.1\% & 3.3$\times$ \\
fashion-mnist   &  \textbf{86.4} & 0.0  &  \textbf{86.4} & \textbf{46.9} & +46.9\% & $\infty$\phantom{$\times$} \\
ionosphere      &           96.7 & 0.0  &  \textbf{97.8} & \textbf{84.4} & +84.4\% & $\infty$\phantom{$\times$} \\
mnist           &           94.9 & 0.0  &  \textbf{95.6} & \textbf{81.7} & +81.7\% & $\infty$\phantom{$\times$} \\
mnist-1-5       &  \textbf{99.8} & 19.0 &           97.9 & \textbf{94.0} & +75.0\% & 4.9$\times$ \\
mnist-2-6       &  \textbf{99.2} & 0.0  &           98.1 & \textbf{88.7} & +88.7\% & $\infty$\phantom{$\times$} \\
sensorless      &  \textbf{99.9} & 22.2 &           94.7 & \textbf{57.4} & +35.2\% & 2.6$\times$ \\
wine            &  \textbf{94.0} & 62.0 &           92.0 & \textbf{68.0} & +6.0\% & 1.1$\times$ \\
\hline
\textbf{Average} &  \textbf{94.7} & 19.8 & 92.9 & \textbf{74.0} & +54.3\% & 3.4$\times$ \\
\hline
\end{tabular}
}

\medskip
\resizebox{0.9\columnwidth}{!}{
\begin{tabular}{| l | cc | cc | cc|}
\hline
\multirow{2}{*}{\textbf{Dataset}}& \multicolumn{2}{c |}{$\textit{leaves}(C)$} & \multicolumn{2}{c |}{$\textsc{Eff}_{\textit{acc}}(C)$} & \multicolumn{2}{c |}{$\textsc{Eff}_{\textit{st}}(C)$} \\
\cline{2-7}
 & \textbf{RF} &  \textbf{\our{}} & \textbf{RF} &  \textbf{\our{}} & \textbf{RF} &  \textbf{\our{}} \\
\hline
breast-cancer    &   641 &      4 & \fn{1.56}{-3} & \fn{2.50}{-1} & \fn{1.59}{-4} & \fn{2.21}{-1} \\
cod-rna          & 89757 &     85 & \fn{1.09}{-5} & \fn{1.13}{-2} & \fn{6.94}{-6} & \fn{1.06}{-2} \\
collision-det.\! & 39678 &     96 & \fn{2.39}{-5} & \fn{9.11}{-3} & \fn{5.29}{-6} & \fn{4.73}{-3} \\
diabetes         & 2583  &     83 & \fn{3.04}{-4} & \fn{9.15}{-3} & \fn{8.04}{-5} & \fn{8.29}{-3} \\
fashion-mnist    & 47549 & 119986 & \fn{1.81}{-3} & \fn{7.20}{-4} & \fn{0.00}{+0} & \fn{3.90}{-4} \\
ionosphere       & 493   &     17 & \fn{1.96}{-3} & \fn{5.75}{-2} & \fn{0.00}{+0} & \fn{4.97}{-2} \\
mnist            & 45268 & 133652 & \fn{2.09}{-3} & \fn{7.15}{-4} & \fn{0.00}{+0} & \fn{6.11}{-4} \\
mnist-1-5        & 3231  &     30 & \fn{3.08}{-2} & \fn{3.26}{+0} & \fn{5.88}{-3} & \fn{3.13}{+0} \\
mnist-2-6        & 4881  &     76 & \fn{2.03}{-2} & \fn{1.29}{+0} & \fn{0.00}{+0} & \fn{1.17}{+0} \\
sensorless       & 15704 &    150 & \fn{6.36}{-5} & \fn{6.31}{-3} & \fn{1.41}{-5} & \fn{3.83}{-3} \\
wine             & 220   &     15 & \fn{4.27}{-3} & \fn{6.13}{-2} & \fn{2.82}{-3} & \fn{4.53}{-2} \\
\hline\end{tabular}
}
 \caption{Comparison between RF and \our{}.}
 \label{tab:vs-rf}
\end{figure}

\begin{figure*}[t]
{\centering
\includegraphics[scale=0.545]{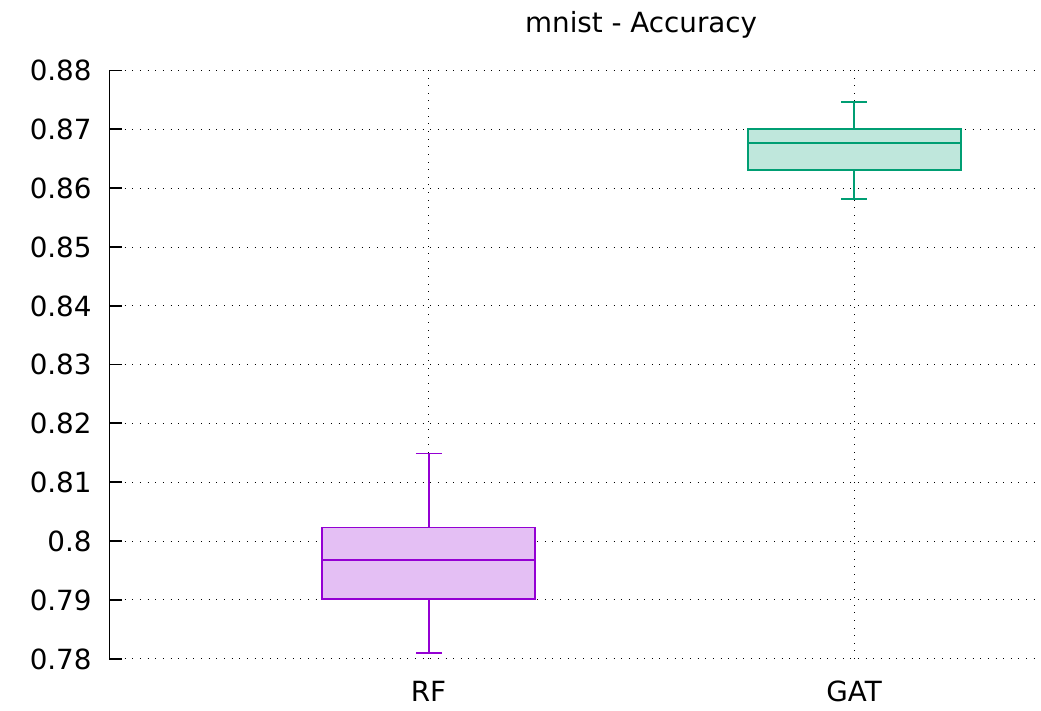}
~~
\includegraphics[scale=0.545]{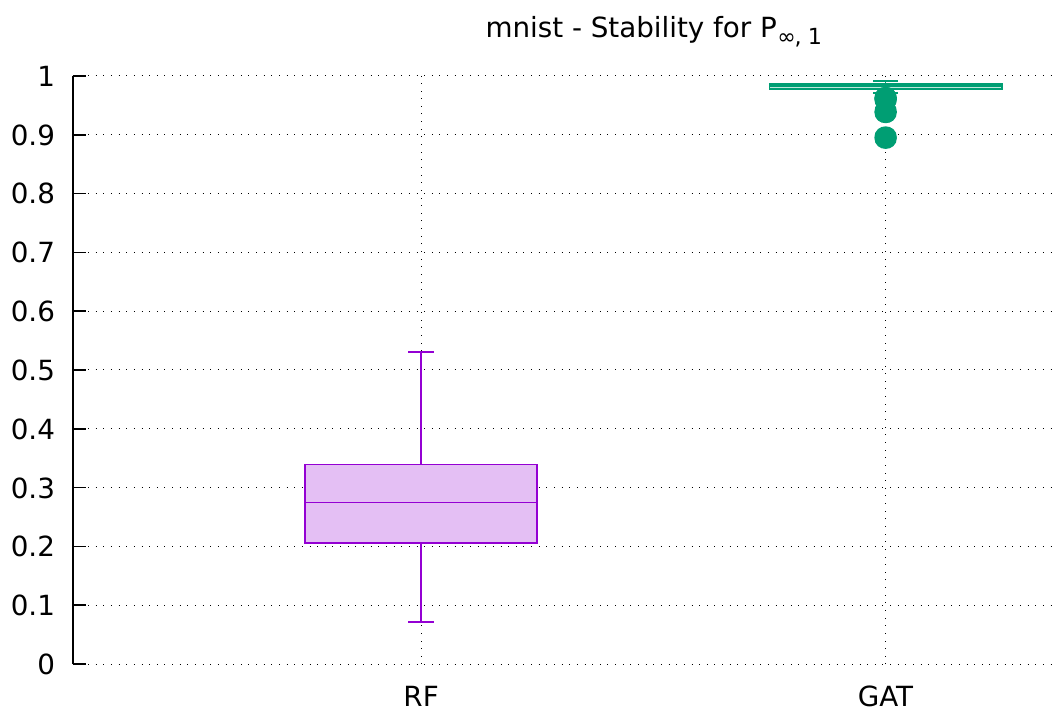}\par
}
\caption{Box plots for mnist accuracy (left) and stability (right) of RF vs.\ \our{}.}
 \label{fig:boxplot}
\end{figure*}

It turns out that all the \our{} models are significantly more stable than RFs ($+54.3\%$ on average). On the other hand, the average accuracy slightly decreased ($-1.8\%$) w.r.t.\ natural RFs. We also emphasize that, with respect to Fig. \ref{fig:variance}, every single \our{} tree outperforms the corresponding Random Forest in terms of stability on every dataset but wine whose RF model already features a significant stability >60\%. With a very significant rise in stability at least a slight drop in accuracy is generally expected, although it is worth observing that in some notable cases the accuracy increased (mnist) or remained the same (fashion-mnist). In Fig.~\ref{fig:boxplot} we display two box plot diagrams which compare accuracy and stability for $P_{\infty,1}$, given as values in $[0,1]$, of all the 50 decision trees composing the RF and \our{} models trained on mnist. The boxes are drawn from first to third quartiles, whose horizontal lines denote the median, and the bounds of the vertical interval denote the lowest and highest data points. 
Interestingly, it can be observed  that each single robust decision tree trained by \our{} is significantly more accurate and stable than each tree of the RF. We also compare an  \emph{efficiency metric} of RFs and \ours{}, where the efficiency of a classifier $C$ w.r.t.\ a metric $m$ measures which size of $C$ is required to achieve a given value of $m$. Here, we define an efficiency metric for a tree ensemble $C$ which takes into account how many leaves of $C$ are needed to reach a given performance of accuracy and stability: if $m(C)\in [0,1]$ is the performance of $C$ for a metric $m$ then the corresponding efficiency of $C$ is defined by $\textsc{Eff}_{m}(C) \ud m(C)/\textit{leaves}(C)$, where $\textit{leaves}(C)\ud\textstyle\sum_{t\in C} \textit{leaves}(t)$ is the total number of leaves of trees in $C$. Thus, the higher $\textsc{Eff}_{m}(C)$ the better is the efficiency of $C$ for $m$. 
The second table in  Figure~\ref{tab:vs-rf} compares the number of leaves and efficiency for accuracy and stability (w.r.t.\ $P_{\infty, \epsilon}$) for RFs and \ours{}. It turns out that \ours{} are much more efficient than RFs: the average of the ratios $\frac{\textit{leaves}(\our{})}{\textit{leaves}(\text{RF})}$ is $51.4\%$ while the average relative efficiency gains of \ours{} w.r.t.\ RFs  (i.e., the ratio $\textsc{Eff}_{m} (\our{})/\textsc{Eff}_{m} (\text{RF})$)
are, resp., 174$\times$ for accuracy, and 1825$\times$ for stability (by excluding in this average 4 datasets whose $\textsc{Eff}_{\textit{st}}$ for RFs  is close to 0).

\begin{figure}[t]
\centering
\resizebox{0.8\columnwidth}{!}{
\begin{tabular}{| l | r r r | r r r |}
\hline
\multirow{3}{*}{\textbf{Dataset}} & \multicolumn{3}{c |}{\textbf{CZBH19}} & \multicolumn{3}{c |}{\textbf{\our{}}} 
\\
\cline{2-7}
  & \multirow{2}{*}{\textit{acc}\%} & \multirow{2}{*}{\textit{rob}\%} & \multirow{2}{*}{obj.~$\varphi_T$} & \multirow{2}{*}{\textit{acc}\%} & \multirow{2}{*}{\textit{rob}\%} & \multirow{2}{*}{obj.~$\varphi_T$} \\
  & & & & & & \\
\hline
breast-cancer\! &  99.3 & 86.9 & 0.98 & \textbf{100.0} & \textbf{89.1} & \textbf{0.99}  \\
cod-rna &  89.8 & 75.8 & 0.88 & \textbf{95.6} & \textbf{88.3} & \textbf{0.95}  \\
diabetes &  \textbf{77.9} & \textbf{59.7} & \textbf{0.76} & 76.0 & 59.1 & 0.74 \\
fashion-mnist\! &  85.6 & 34.9 & 0.80 & \textbf{86.4} & \textbf{44.8} & \textbf{0.82}  \\
mnist-1-5 &  \textbf{99.7} & \textbf{97.1} & \textbf{0.99} & 97.9 & 93.1 & 0.97  \\
mnist-2-6 &  \textbf{99.5} & \textbf{93.1} & \textbf{0.99} & 98.1 & 88.1 & 0.97 \\
\hline
\textbf{Average} & 91.9 & 74.6 & 0.90 & \textbf{92.3} & \textbf{77.1} & \textbf{0.91}  \\ \hline
\end{tabular}
}

\medskip
\resizebox{0.8\columnwidth}{!}{
\begin{tabular}{| l | r r r  | r r r  |}
\hline
\multirow{2}{*}{\textbf{Dataset}} & \multicolumn{3}{c |}{\textbf{AH19}} & \multicolumn{3}{c |}{\textbf{\our{}}}  \\
  & \textit{acc}\% & \textit{rob}\% & obj.~$\varphi_T$ &  
     \textit{acc}\% & \textit{rob}\% & obj.~$\varphi_T$ 
     \\
\hline
breast-cancer\! &  99.3 & \textbf{93.4} & \textbf{0.99} &  \textbf{100.0} & 89.1 & \textbf{0.99}  \\
cod-rna &  93.1 & 78.7 & 0.92 &  \textbf{95.6} & \textbf{88.3} & \textbf{0.95}\\
diabetes &  72.7 & \textbf{64.3} & 0.72 &\textbf{76.0} & 59.1 & \textbf{0.74} \\
fashion-mnist\! &  85.8 & \textbf{76.8} & \textbf{0.85} &  \textbf{86.4} & 44.8 & 0.82 \\
mnist-1-5 &  \textbf{99.8} & \textbf{98.7} & \textbf{0.99} & 97.9 & 93.1 & 0.97 \\
mnist-2-6 &  \textbf{99.3} & \textbf{96.2} & \textbf{0.99} & 98.1 & 88.1 & 0.97 \\
\hline
\textbf{Average} & 91.7 & \textbf{84.7} & \textbf{0.91} &  \textbf{92.3} & 77.1 & \textbf{0.91}  \\ \hline
\end{tabular}
}

\medskip
\resizebox{0.8\columnwidth}{!}{
\begin{tabular}{| l | c | c | r r r |}
\hline
\multirow{3}{*}{\textbf{Dataset}} & 
\multirow{2}{*}{\textbf{AH19}} & \multirow{2}{*}{\textbf{\our{}}} 
&\multicolumn{3}{c |}{\textbf{Efficiency Gain}}
\\
  &  &  & \multicolumn{3}{c |}{\textbf{\our{}/AH19}}\\
\cline{2-6}
& {\#\,\textit{leaves}} & {\#\,\textit{leaves}} &\textit{leaves} & \textit{acc} & \textit{rob} \\
\hline
breast-cancer\! &    80 &      4 & \textbf{5.0}\% & \textbf{20.1}$\times$  & \textbf{18.9}$\times$ \\
cod-rna         &   913 &     85 & \textbf{9.3}\% & \textbf{11.0}$\times$  & \textbf{12.1}$\times$ \\
diabetes        &    79 &     83 & 105.0\%        &           1.0$\times$  & 0.9$\times$ \\
fashion-mnist\! & 97279 & 119986 & 123.3\%        &           0.8$\times$  & 0.5$\times$ \\
mnist-1-5       &  3258 &     30 & \textbf{0.9}\% & \textbf{106.5}$\times$ & \textbf{102.4}$\times$ \\
mnist-2-6       &  3777 &     76 & \textbf{2.0}\% & \textbf{49.1}$\times$  & \textbf{45.5}$\times$ \\
\hline
\end{tabular}
}
\caption{Comparison between \our{} and AH19, CZBH19.}
 \label{comparison-table}
\end{figure}

\subsection{\our{} vs Robust Gradient Boosted Trees}
Finally, we compare our \our{} robust models with the adversarially trained tree models of \cite{Andriushchenko19}, denoted by AH19, and \cite{ChenZBH19}, denoted by CZBH19. These are gradient boosted decision trees of the same type of XGBoost trees \cite{xgboost}, which, to the best of our knowledge, represent the state-of-the-art of adversarially trained GBDTs. Although our \our{} robust training generates tree ensembles which are random forests, and RFs and GBDTs are tree ensemble models with unrelated training principles, we nevertheless compare these different models since these robust GBDTs were the only adversarially trained decision trees found in literature. We considered 6 common datasets and, as already recalled, the perturbation  $P_{\infty, \epsilon}$ is exactly the same used in \cite{Andriushchenko19} and \cite{ChenZBH19}. Let us remark that the accuracy and robustness metrics of AH19 and CZBH19 models are taken from Table~7 of the supplemental of \cite{Andriushchenko19}, because silva cannot be used to compute the robustness of GBDTs.  We also compare the objective performance function $\varphi_T$ used for training \our{}s, where accuracy and robustness weigh, resp., 90\% and 10\%.

The results of this comparison are summarized by the tables in 
Figure~\ref{comparison-table}.  
For AH19 models, we also compare their size, i.e.\ total number of leaves, and  their efficiency with \our{} models. The third table in Figure~\ref{comparison-table} displays the relative efficiency gains of \our{} w.r.t.\ AH19 models, which are, resp., the ratios $\frac{\textit{leaves}(\our{})}{\textit{leaves}(\text{AH19})}$, $\frac{\textsc{Eff}_{\textit{acc}} (\our{})}{\textsc{Eff}_{\textit{acc}} (\text{AH19})}$ and $\frac{\textsc{Eff}_{\textit{rob}} (\our{})}{\textsc{Eff}_{\textit{rob}} (\text{AH19})}$.

On average, it turns out that: 
\begin{itemize}
\item[(i)] AH19 models are moderately more robust ($+7.6\%$) than \ours{}; 
\item[(ii)] \ours{} are slightly more robust ($+2.5\%$) than CZBH19 models; 
\item[(iii)] \our{} models are slightly more accurate of both AH19 ($+0.6\%$) and CZBH19 ($+0.4\%$) models; 
\item[(iv)] all three models exhibit the same average performance according to the objective function $\varphi_T$; 
\item[(v)] \our{} models are significantly more efficient than AH19 models both for size (average $40.9\%$), accuracy (average $31.4\times$) and robustness (average $30.1\times$).
\end{itemize}
Our \our{} models compete with and often improve on the robust GBDTs of \cite{Andriushchenko19,ChenZBH19} while being more compact and therefore interpretable and efficient tree models.

\section{Conclusion}
\label{sec:conclusion}
We believe that this work contributes to push forward the use of formal verification methods in machine learning, in particular a very well known program analysis technique such as abstract interpretation has been proved successful for training decision tree classifiers which are both accurate and robust and compete with and often improve on the state-of-the-art of adversarial training of gradient boosted decision trees while being much more compact and therefore interpretable and efficient tree models. As future work, we plan to investigate the problem of fairness verification \cite{Chouldechova,mehrabi2019survey}  for ensembles of decision trees by leveraging abstract interpretation techniques along the lines of this paper. The final goal will be to design a fairness-aware learning algorithm for decision trees, for both notions of individual and group fairness \cite{Binns}.

\paragraph{\textbf{\textit{Acknowledgments.}}}
This work has been partially funded by the University of Padova, under the SID2018 project ``Analysis of STatic Analyses (ASTA)'', by the Italian Ministry of Research MIUR, under the PRIN2017 project no. 201784YSZ5 ``AnalysiS of PRogram Analyses (ASPRA)''.
The research of Fran\-ce\-sco Ranzato has been partially funded 
by Facebook inc.\ under the Probability and Programming Research Award ``Adversarial Machine Learning by Morphological
Abstract Interpretation''.
The research of Marco Zanella has been funded by Fondazione Bruno Kessler (FBK), Trento, Italy. 


\end{document}